# High-Fidelity 3D Lung CT Synthesis in ARDS Swine Models Using Score-Based 3D Residual Diffusion Models

Siyeop Yoon[1], Yujin Oh[1], Xiang Li[1], Yi Xin[2], Maurizio Cereda[2], and Quanzheng Li[1,*]
[1] Center for Advanced Medical Computing and Analysis, Department of Radiology, and
[2] Department of Anesthesiology, Critical Care, and Pain, Medicine Massachusetts General Hospital and Havard Medical School, 55 Fruit Street, Boston, MA, USA 02114;

*Corresponding Author, li.quanzheng@mgh.harvard.edu

## ABSTRACT

Acute respiratory distress syndrome (ARDS) is a severe condition characterized by lung inflammation and respiratory failure, with a high mortality rate of approximately 40%. Traditional imaging methods, such as chest X-rays, provide only two-dimensional views, limiting their effectiveness in fully assessing lung pathology. Three-dimensional (3D) computed tomography (CT) offers a more comprehensive visualization, enabling detailed analysis of lung aeration, atelectasis, and the effects of therapeutic interventions. However, the routine use of CT in ARDS management is constrained by practical challenges and risks associated with transporting critically ill patients to remote scanners. In this study, we synthesize high-fidelity 3D lung CT from 2D generated X-ray images with associated physiological parameters using a score-based 3D residual diffusion model. Our preliminary results demonstrate that this approach can produce high-quality 3D CT images that are validated with ground truth, offering a promising solution for enhancing ARDS management.

## 1. INTRODUCTION

Acute respiratory distress syndrome (ARDS) is a critical and life-threatening condition characterized by severe inflammation and injury to the lungs, often leading to respiratory failure[1]. Despite advancements in respiratory care, ARDS remains associated with a high mortality rate of approximately 40%. Traditional bedside imaging techniques, such as chest X-rays, are commonly used in the clinical management of ARDS but are limited in their ability to provide a comprehensive view of the lungs, as they only offer two-dimensional representations. In contrast, three-dimensional (3D) computed tomography (CT) imaging has proven to be a powerful tool in visualizing the complex and heterogeneous nature of lung pathology in ARDS[2,3]. 3D CT allows for a detailed assessment of lung aeration, atelectasis, and the effects of various therapeutic interventions, such as increasing positive end-expiratory pressure (PEEP)[4]. The ability to 3D visualization and quantification provides critical insights that can significantly impact the management and of ARDS. However, the routine use of 3D CT imaging in ARDS management is limited by practical constraints, such as the need for frequent imaging, radiation exposure, and the risks associated with transporting critically ill patients to the imaging facility.

Recent advancements in deep learning have led to the development of methods for 3D CT reconstruction and synthesis that require lower radiation doses, fewer imaging views, or limited angles[5–7]. Generative AI models, such as diffusion models, have gained attention for their superior performance in generating realistic images, offering stable training, and improved generalizability. Despite these advances, ultra-sparse view CT reconstruction remains a highly challenging and ill-posed problem. The generated CT images often exhibit significant variability and carry the risk of producing "hallucinations."[8] Although these hallucinations may appear realistic, they can lack clinical accuracy and relevance, making them potentially unsuitable for ARDS patients, who require precise and detailed assessments of lung aeration and other critical parameters.

In this work, we explore the potential of score-based 3D residual diffusion models to synthesize high-fidelity 3D lung CT volumes by utilizing prior 3D CT data alongside limited 2D imaging data and various physiological parameters, such as respiratory phase, PEEP, and tidal volumes. The 3D residual diffusion model leverages rich prior CT information to synthesize 3D CT images that are accurately aligned with the target respiratory phase and physiological parameters. The proposed model was trained and tested on ARDS swine models by comparing its assessments of lung volumes (LV) and normal-aerated lung. Our preliminary results demonstrate the potential of the 3D residual diffusion model to produce high-quality 3D CT images from minimal input data, reducing the need for repeated CT scans while still providing the critical information necessary for effective treatment planning.

## 2. METHODS

### 2.1 Score-based Diffusion Model for Generating Residuals of Lung CT

In this section, we adapt the conditional score-based diffusion model framework to synthesize the Residual of CTs from prior CT data and physiological conditions, including bedside X-ray images. Residual CT is defined as the difference between the target CT (the CT scan we aim to generate) and the prior CT (the CT scan already available). Residual of CT, $x_{Residual} = x_{trg} - x_{prior}$, captures the changes between the prior and target CTs.

Initially, consider a set of distributions, $p(x_{Residual}; \sigma, x_{prior})$, obtained by adding a Gaussian noise with a standard deviation $\sigma$ to the residual of CT images conditioned on priors CT and physiological conditions. As $\sigma$ increases, the noise progressively overwhelms the residuals, and at a maximum noise level $\sigma_{max}$, the distribution $p(x_{Residual}; \sigma_{max}, x_{prior})$ almost indistinguishable from pure Gaussian noise. The transition from a distribution dominated by the residuals to one dominated by noise can be described using a probability flow ordinary differential equation (ODE) for both the forward process (as $\sigma$ increases) and the backward process (as $\sigma$ decreases)[9]. The probability flow ODE is expressed as:

$$dx = -\dot{\sigma}(t)\sigma(t)\nabla_x \log p(x_{Residual}; \sigma(t), x_{prior}) \, dt \quad (1)$$

, where the $\nabla_x \log p(x_{Residual}; \sigma(t), x_{prior})$ represent the score function, the gradient of log distribution of residuals $x_{Residual}$, guiding the reverse diffusion process toward regions of higher probability density at a given noise level $\sigma(t)$ and prior $x_{prior}$. To improve the model's adaptability across varying noise levels $\sigma(t)$, the score function with scaling factor dependent on $\sigma$ as:

$$\nabla_x \log p(x_{residual}; \sigma(t), x_{prior}) = \frac{D_\theta(x_{residual}; \sigma(t), x_{prior}) - x_{residual}}{\sigma(t)} \quad (2)$$

, where $D_\theta$ is a denoising score-matching 3D U-net parameterized by $\theta$. Score matching in the diffusion process can be done by training the neural network to minimize the following loss that aims to predict the $x_{residual}$ from its noisy version

$$L = \mathbb{E}_{x_{residual}} \mathbb{E}_{n \sim N(0,\sigma^2 I)} \| D_\theta(x_{residual} + n; \sigma, x_{prior}) - x_{residual} \|. \quad (3)$$

Once the denoising score-matching network is trained, we use the trained model to sample and estimate the $x_{residual}$ in the reverse process using Equation (1) with EDM stochastic reverse diffusion sampler. The prior CT, X-ray images, and ventilation parameters were concatenated along the channel dimension to serve as conditioning inputs for the network. Specifically, the orthogonal X-ray images were back-projected to construct the 3D volumetric input. The ventilation parameters were normalized by min-max values and resized to match to the dimensions of the prior CT.

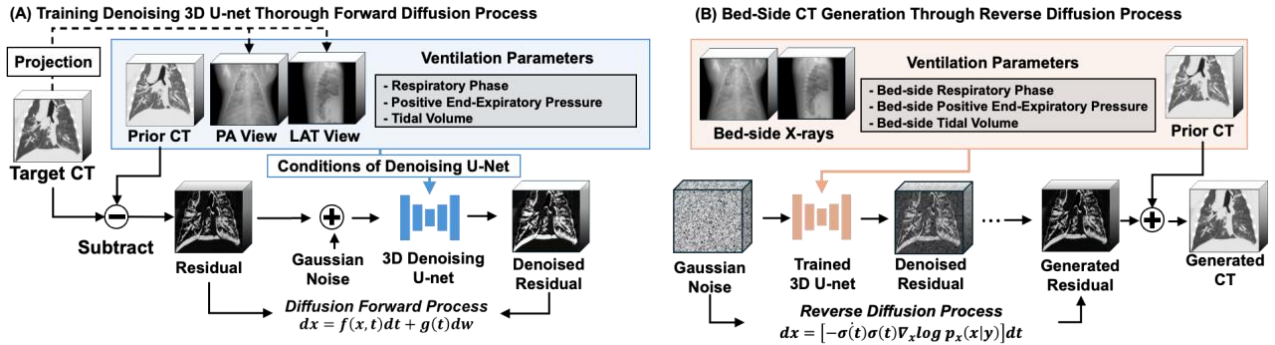

**Figure 1. Overview of the Conditional Score-Based Diffusion Model for Synthesizing Residuals of Lung CT**. The forward diffusion process is performed to train a denoising score-matching 3D U-net that estimate the residual of CTs using network conditions (prior CT data, PA and LAT view X-rays, and ventilation parameters). The reverse diffusion process is performed to generate the Generated CT from the bedside X-rays and target ventilation parameters.

## 2.3 Datasets and Implementation Details

Our model was trained and tested on an ARDS swine model comprising a total of 50 pigs. The dataset was randomly split into a training set (N=40) and a testing set (N=10). To create ARDS model, all animals received 3.5 ml/kg of hydrochloric acid (pH 1.0) while in the supine position. This was administered in 5-ml aliquots via bronchoscopy (Ambu Inc., USA) into the lobar bronchi. The pigs were anesthetized, and muscle relaxation was maintained throughout the imaging procedures. Each pig was ventilated and scanned multiple times under various ventilation parameters, including tidal volume ranging from 6 to 12 ml·kg$^{-1}$, positive end-expiratory pressure (PEEP) ranging from 3 to 15 cm $H_2O$ and different respiratory phases (end-inspiratory and end-expiratory), as well as different imaging positions (supine and prone). Scanning was conducted during both the early stage (30 minutes after hydrochloric acid administration) and the stabilized stage (24 hours after hydrochloric acid administration) of ARDS.

CT scans were acquired using a Siemens SOMATOM Force scanner (Siemens Medical Systems, Germany) with a spatial resolution of 0.53×0.53×0.5 mm³, which were then resampled to 2×2×2 mm³ using linear interpolation. The training dataset consisted of 701 CT scans from 40 pigs, while 280 scans were available for the testing dataset from 10 pigs. To create the residual CTs, CT scans obtained from the same ARDS phase and imaging position were randomly paired, differing only in ventilation parameters and respiratory phases. Orthogonal digitally reconstructed radiographs (DRRs) were generated from the target CTs to serve as the X-ray images used in this study.

We implemented our 3D residual diffusion model on the original EDM framework, modifying the network architecture to include 3D convolutional layers. For training, we randomly sampled patches of size 128³. The model was fed patches containing residuals, prior CT, X-rays, ventilation parameters, respiratory phase, and patch coordinates. We set the number of channels to 64 and used mixed precision. The training was conducted using the Adam optimizer with a learning rate of 0.0005 and a batch size of 24, running on 8 NVIDIA A100 40GB GPUs for 5 days, completing 200k gradient updates. We kept the same hyperparameters as in the original EDM[9], unless otherwise specified. The final generated CT were synthesized by adding the estimated residual CT from reverse diffusion process (the number of function evaluation=200).

## 2.4 Evaluation

We aimed to investigate the feasibility of quantifying LV and normally-aerated lung using the proposed diffusion model. For LV measurements, we employed an open-source deep learning model for lung segmentation[10]. Normally-aerated lung was defined as areas with Hounsfield units in the range of [-900, -500] with in the lung mask[11]. The normally-aerated lung was reported as the percentage of Normally-aerated Lung Volume in Total Lung Volume. To evaluate the measurement agreement, we used Bland-Altman analysis, linear regression, and Pearson correlation coefficient.

## 3. RESULTS

### 3.1 Comparing the original lung CT and generated CT in Lung Volume assessment.

The assessment of LVs were performed at the end of inspiration and expiration, where the generated CT was synthesized using the opposite phase CT as prior data, combined with target phase X-ray images and ventilation parameters. Bland-Altman analyses (Fig 2A) showed good agreement between the original and generated CTs for LV. Bias and 95% limits of agreement for LV were as follows: end-inspiratory of 3 ml [−67, 73ml] and end-expiratory of 30 ml [-38, 100ml]. In the linear regression (Fig 2B), LVs of the generated CTs strongly correlated with the original CTs (All $R^2$ >0.98).

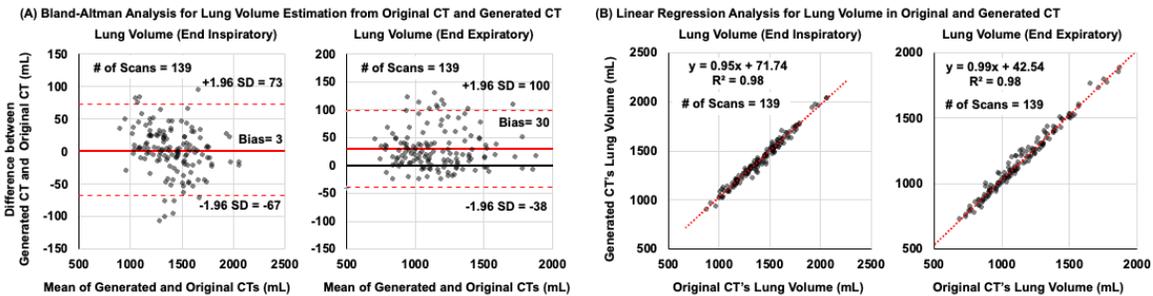

**Figure 2. Evaluation of Lung Volume Quantification Using the Proposed 3D Residual Diffusion Model.** (A) Bland Altman analysis, the solid red line and red dotted lines represent mean differences (Bias) and limits of agreement (±1.96 SDs) of lung volume assessment. (B) In the linear regression analysis, the correlation coefficient ($R^2$) and slope (y) are shown, and a positive correlation was observed for all parameters.

**3.2 Comparing the original lung CT and generated CT in Normally-Aerated Lung assessment at the target PEEP.**

To evaluate the feasibility of the proposed 3D residual diffusion model in assessing the normally-aerated lung rate associated with changes in PEEP levels, we generated CT images at PEEP levels of 8 and 13 cm $H_2O$, starting from a baseline PEEP of 3 cm $H_2O$. These generated CT images were then used to assess and quantify changes in lung aeration as PEEP was adjusted. Bland-Altman analyses indicated good agreement between the original and generated CT images regarding lung aeration at increased PEEP levels, with mean differences in normally-aerated lung rate of -1% [−8%, 6%]. Additionally, the normally-aerated lung rate of the generated CT showed a strong correlation with the original ($R^2$ >0.91). The representative CT images further demonstrate that the generated CTs accurately reflect changes in normally-aerated lung regions when compared to both prior and ground truth CT images.

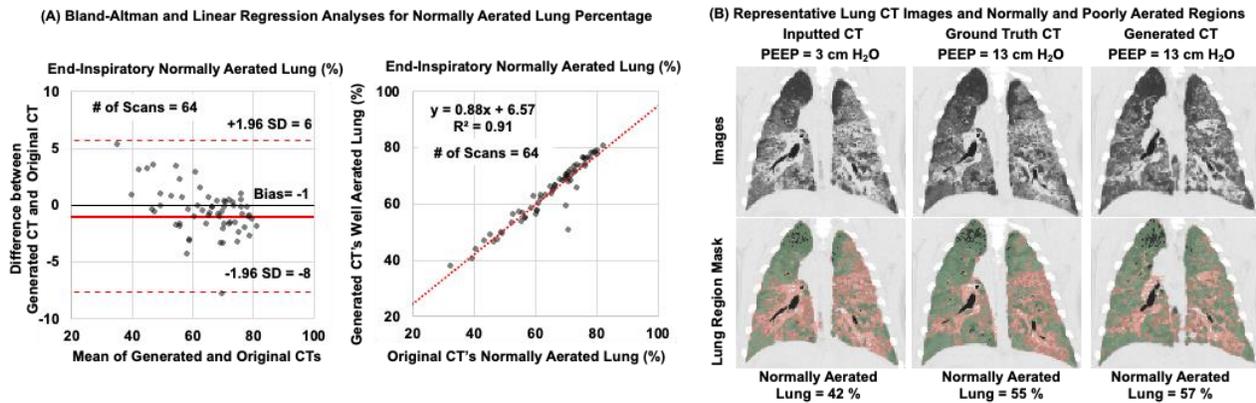

**Figure 3. Evaluation of the 3D residual diffusion model for assessing normally-aerated lung with changes in PEEP levels.** (A) Bland-Altman and linear regression analysis show good agreement between original and generated CT images for lung aeration and strong correlation ($R^2$ >0.91). (B) Representative CT images illustrate that the generated CTs accurately reflect changes in normally-aerated lung regions compared to prior and ground truth CTs. The normally-aerated and poorly-aerated lung regions were colored green and red, respectively.

## 4. CONCLUSION AND DISCUSSION

In this study, we explored the feasibility of using a score-based 3D residual diffusion model to synthesize high-fidelity 3D lung CT volumes for ARDS management. Our preliminary evaluation in the ARDS pigs indicate that the model can accurately generate 3D CT that reflect changes in lung volume and aeration in response to varying respiratory phases and PEEP levels, demonstrating strong agreement with original CT and high correlation coefficients. By leveraging prior CT, limited 2D imaging, and physiological parameters, our approach offers a potential solution to challenges like radiation exposure and logistical difficulties in critically ill patients. These results suggest that the proposed 3D residual diffusion model's potential for reducing the need for repeated CT scans while providing 3D quantification for ARDS treatment.

Despite these promising results, several factors need to be considered. The study's reliance on a swine model may limit the direct applicability of the findings to human patients. Additionally, the inherent limitations of generative models, such as the potential for "hallucinations," must be addressed through rigorous validation in diverse clinical settings. Future works are warranted on refining the model, extending its applicability to other respiratory conditions, and validating its performance in human subjects. In conclusion, the 3D residual diffusion model presents a promising advancement in utilizing AI to support the management of ARDS and other complex pulmonary conditions.

## ACKNOWLEDGEMENTS

No potential conflicts of interest relevant to this study exist.